\documentclass[12pt,a4paper]{article}

\usepackage[british]{babel}

\usepackage[a4paper,top=2cm,bottom=2cm,left=2.5cm,right=2.5cm,marginparwidth=1.75cm]{geometry}

\usepackage[style=apa, backend=biber]{biblatex} 
\usepackage{amsmath}
\usepackage{graphicx}
\usepackage[colorlinks=true, allcolors=blue]{hyperref}
\usepackage{hyperref}
\usepackage[title]{appendix}
\usepackage{mathrsfs}
\usepackage{amsfonts}
\usepackage{booktabs} 
\usepackage{caption}  
\usepackage{threeparttable} 
\usepackage{algorithm}
\usepackage{algorithmicx}
\usepackage{algpseudocode}
\usepackage{listings}
\usepackage{enumitem}
\usepackage{chngcntr}
\usepackage{booktabs}
\usepackage{lipsum}
\usepackage{subcaption}
\usepackage{authblk}
\usepackage[T1]{fontenc}    
\usepackage{csquotes}       
\usepackage{diagbox}

\usepackage{setspace}
\usepackage{titlesec}
\titleformat{\section} 
  {\normalfont\Large\bfseries}{\thesection.}{1em}{}
  
\usepackage{lineno} 

\rightlinenumbers 

\usepackage{float}   
\usepackage{caption} 
\title{More Consideration for the Perceptron}

\author[1]{Slimane LARABI}
\affil[1]{\small Computer Science Faculty, USTHB University, BP 32, EL ALIA, 16111, Algiers, Algeria}
\affil[*]{Corresponding author: \texttt{slarabi@usthb.dz}}

\date{}  

\begin{document}
\maketitle

\begin{abstract}
In this paper, we introduce the gated perceptron, an enhancement of the conventional perceptron, which incorporates an additional input computed as the product of the existing inputs. This allows the perceptron to capture non-linear interactions between features, significantly improving its ability to classify and regress on complex datasets. We explore its application in both linear and non-linear regression tasks using the Iris dataset, as well as binary and multi-class classification problems, including the PIMA Indian dataset and Breast Cancer Wisconsin dataset. Our results demonstrate that the gated perceptron can generate more distinct decision regions compared to traditional perceptrons, enhancing its classification capabilities, particularly in handling non-linear data. Performance comparisons show that the gated perceptron competes with state-of-the-art classifiers while maintaining a simple architecture.
\end{abstract}

\textbf{Keywords}: Gated Perceptron, Arithmetic Gate AND, Non linearity, Non linear Regression.

\section{Introduction}

The first artificial neuron was introduced by Warren McCulloch in 1943 [\cite{1}]. In this model, without any training, the weighted sum of inputs is compared to a threshold to determine the neuron's output. In the 1950s, Frank Rosenblatt proposed a learning rule for training neural networks, introducing the concept of the perceptron [\cite{2}]. However, the limitations of perceptrons, particularly their inability to handle non linearity, were highlighted by Marvin Minsky and Seymour Papert [\cite{3}]. They demonstrated that perceptrons could not account for nonlinear relationships.
Subsequently, the development of multilayer perceptrons and training algorithms like back propagation [\cite{6}] enabled the processing of nonlinear problems. 

Using only one neuron in a single-layer neural network for binary classification is equivalent to a simple linear classifier. This approach can work well if the data is linearly separable (by a straight line or hyperplane in higher dimensions). However, if the data is more complex and not linearly separable, using just one neuron in a single layer might not yield good results. With high number of features, the data might have complex interactions, which a single neuron won’t be able to capture.

The core idea being proposed is that the addition of an AND gate allows for the introduction of an additional input, effectively enabling the perceptron to capture nonlinearity in data. This is a significant departure from the conventional perceptron, which struggles to classify nonlinear data, leading researchers historically to rely on more sophisticated methods such as Support Vector Machines (SVM), Linear Discriminant Analysis (LDA), k-Nearest Neighbors (k-NN), and various ensemble methods like Random Forests and Gradient Boosting Machines (GBM).

In this paper, we aim to explore the utility of the gated perceptron in the context of classification tasks, especially as an alternative to more complex architectures and algorithms that are typically used when dealing with data that exhibits high dimensionality or nonlinearity.

The paper is organized as follows. In Section 2, we define the gated perceptron and present its properties. Section 3 is devoted to the application of the gated perceptron for computing linear and nonlinear regression. We explain in Section 4 how to apply the gated perceptron to solve binary and multi-class classification problems. Experiments conducted on three common datasets are presented and compared to the state-of-the-art. Finally, we conclude with a discussion on generalizing the gated perceptron to more complex data and outline potential directions for future research.

\section{The Gated Perceptron and Proprieties}

\begin{figure}[!ht]
\centering
\includegraphics[width=8cm]{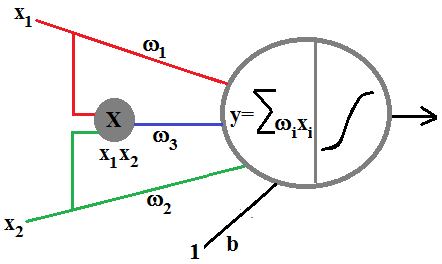}
\caption{\label{fig1} A gated perceptron with two inputs.}
\end{figure}

We define a gated perceptron as a conventional perceptron with inputs, activation function and output, and in addition a new input computed as the product of all inputs. Figure \ref{fig1} shows a gated perceptron with two inputs, $(X_1)$ and $(X_2)$, a third input is generated from these two inputs equal to $X_1*X_2$.

Similar to the conventional perceptron, to each input is assigned a weight, and the weighted sum is calculated as follows:

\begin{equation}\label{eq0}
y= \omega_1 X_1 + \omega_2 X_2 + \omega_3 X_1 X_2+b   
\end{equation}\label{eq01}

In order to study the sum function $y$, we draw its boundary expressed by the following equation.

\begin{equation}
X_2(\omega_2 + \omega_3 X_1) + \omega_1 X_1 + b =0 
\end{equation}

\begin{equation}\label{eq1}
X_2= -\frac{\omega_1 X_1 + b}{\omega_3 X_1 + \omega_2}    
\end{equation}

\begin{figure}[!ht]
\centering
\includegraphics[width=8cm]{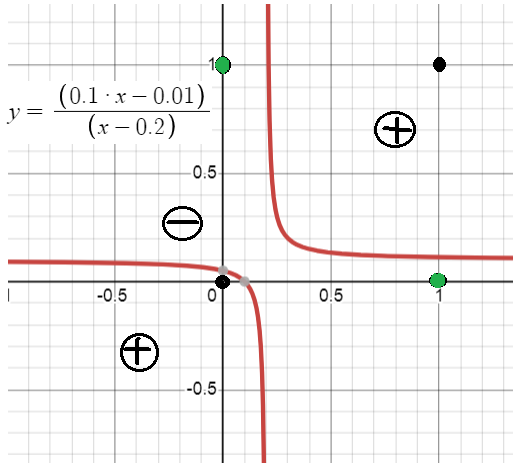}
\caption{\label{fig2} Graphical illustration of the output of the gated perceptron.}
\end{figure}

Figure \ref{fig2} highlights with red color the curved boundary $(y=0)$ dividing the 2D space into three regions with either positive or negative values of $y$, depending on the weights of the expression given in Equation \ref{eq0}. The graphical illustration of the gated perceptron’s output demonstrates its ability to partition the input space into multiple distinct regions, depending on the gate configuration. This flexibility in partitioning is what enables the gated perceptron to handle more complex, non-linear data distributions compared to a traditional perceptron, as shown in Figure \ref{fig2}.

The XOR gate, a classic example of non-linear data, can be solved using a gated perceptron, which finds the corresponding weights as shown in Figure \ref{fig2}. The classification into two regions—negative (including the values (0,1) and (1,0)) and positive (including the values (1,1) and (0,0))—is achieved by the computed weights: $w_1=0.1, w_2=-0.2, w_3=1.0, b=-0.01.$

\begin{figure}[!ht]
\centering
\includegraphics[width=14cm]{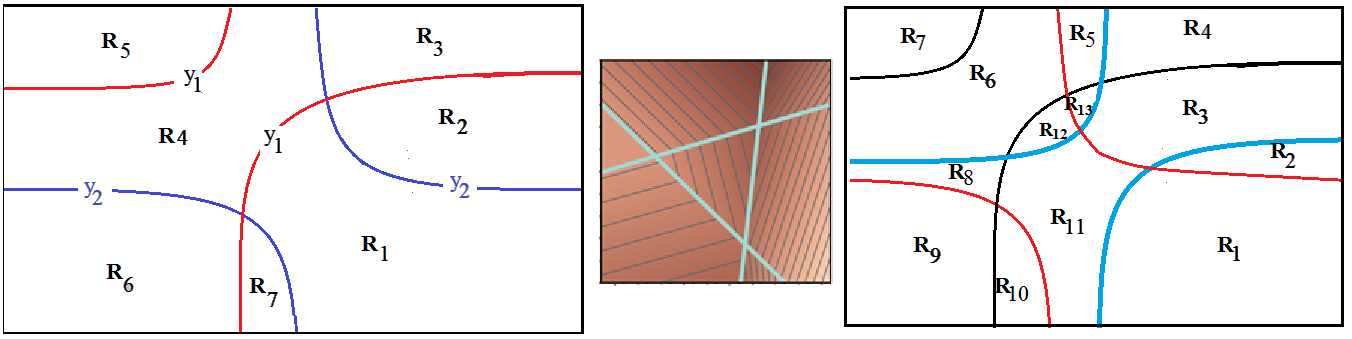}
\caption{\label{fig4} (Left) Graphical illustration of the two outputs $(y_1, y_2)$ of the Sallow Neural Network. (Middle) The 07 regions generated using a shallow neural network with three conventional perceptrons [\cite{4}], (Right) the 13 Regions generated using 3 gated perceptrons}
\end{figure}

Consider a shallow neural network where the input layer consists of two gated perceptrons. The geometric representation of the output from this input layer is shown in Figure \ref{fig4}, which defines seven distinct regions based on the outputs $y_1, y_2$ of the two gated perceptrons. In contrast, a shallow neural network with two inputs and two conventional perceptrons generates only four distinct regions, as explained in [\cite{4}]. Incorporating a third traditional perceptron into the network allows the generation of seven distinct regions. In contrast, adding a third gated perceptron results in 13 distinct regions, as depicted in Figure \ref{fig4}.

\section{The Gated Perceptron for Computing Linear and non Linear Regression}

This section explores the application of the gated perceptron in both linear and non-linear regression tasks. By utilizing gate mechanisms, the perceptron adapts to a wider range of data patterns, allowing it to compute non-linear relationships that traditional perceptrons struggle with. Through the appropriate choice of weights and gate configurations, the gated perceptron demonstrates its capacity to model complex, non-linear functions, as well as simpler, linear relationships.

For the computation of linear regression using a gated perceptron, we consider the Iris dataset [\cite{5}], commonly used in classic regression tasks. This dataset includes four parameters defining the type of plants. To perform regression on this dataset with two classes ('Iris-setosa' and 'Iris-versicolor'), we employ a gated perceptron with two inputs 
$(x_i, x_j)$, where $(i,j = 0..3)$. Figure \ref{fig10} displays the results obtained using one combination of these two parameters; similar results are observed with other combinations. In the figure, green dots represent instances of the first class ('Iris-setosa'), while red dots represent instances of the second class ('Iris-versicolor'). The figures also include regression results obtained using a simple perceptron for comparison.

\begin{figure}[ht!]
\centering
\includegraphics[width=12cm]{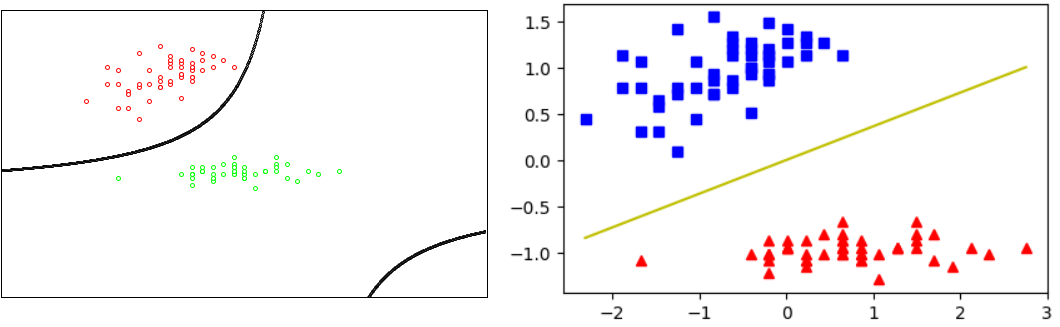}
\caption{\label{fig10} The regression computed using gated (left) and simple (right) perceptron for the parameters $(x_1, x_2)$.}
\end{figure}

The concept of computing the boundaries between three classes of data is based on the loss function $(L)$  defined by equation \ref{eq2p}, where $(x^i_1, x^i_2)$ represent the $i^{th}$ input to the gated perceptron.

\begin{equation}
    L=\sum_{i=1}^{N} l_i =\sum_{i=1}^{N} |class_i-y(x^i_1,x^i_2)|
\end{equation}\label{eq2p}

When tackling nonlinear regression with three classes, we define the following:

- class: The label assigned to each data set, with labels $(+1), (-1)$, and $(+1)$. The labels $(+1)$ corresponds to positive regions, while $(-1)$ corresponds to a negative region.\\
- $l_r$: The learning rate.

The weights $(w_1, w_2, \dots, w_k)$ are updated during training according to equation \ref{eq30}.
\begin{equation}
    \omega_k=\omega_k + lr* \frac{\delta l}{\delta \omega_k}
\end{equation}\label{eq30}

\begin{itemize}
    \item $\omega_1 = \omega_1+lr*((class-y)*x_1$
    \item $\omega_2 = \omega_2+lr*((class-y)*x_2$
    \item $\omega_3 = \omega_3+lr*((class-y)*x_1x_2$
    \item $b = b+lr*(class-y)$.
\end{itemize}

We compute the value of $y$ as described in equation \ref{eq01} and adjust the weights to ensure the output is either positive or negative, depending on the class of the corresponding data point. If the data point belongs to the positive class, we update the parameters until $y$ reaches the target class value, making $y$ negative. Conversely, if the data point belongs to the negative class, we adjust the parameters until $y$ reaches the target class value, making $y$ positive.

\begin{figure}[ht!]
\centering
\includegraphics[width=10cm]{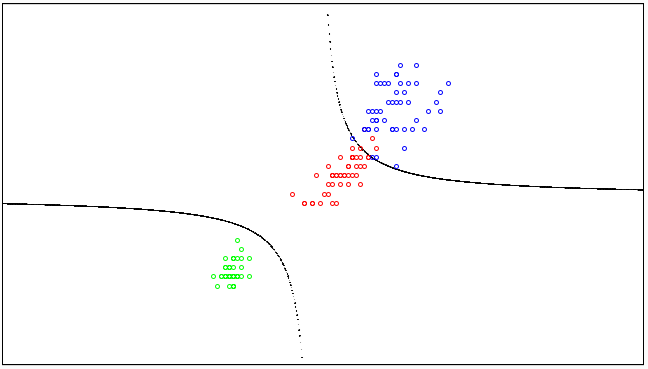}
\caption{\label{fig13} The regression computed using gated perceptron with three classes (iris dataset).}
\end{figure}

Figure \ref{fig13} presents the results of the non-linear regression computation using the two variables $(x_1, x_2)$, which correspond to the third and fourth columns of the Iris dataset [\cite{5}]. The decision boundary is determined with high accuracy. Of the fifty elements in the 'Iris-versicolor' class, three are misclassified, and only one out of fifty elements in the 'Iris-virginica' class is misclassified. All elements of the 'Iris-setosa' class are correctly classified. The learning rate$(lr)$ is set to $(0.05)$ over $40$ epochs.

Finally, we can assert that the gated perceptron offers two key advantages:
\begin{itemize}
    \item Gated perceptrons can generate more distinct regions compared to conventional perceptrons, allowing for finer data separation.
    \item While conventional perceptrons rely on linear boundaries, gated perceptrons use asymptotic boundaries, providing greater flexibility in adjusting region boundaries and enhancing classification performance.
\end{itemize}

\section{The Gated Perceptron for Classification}

In this section, we examine the efficiency of the gated perceptron in solving classification problems. We begin by addressing binary classification, followed by an investigation into its application for multi-class classification.

\subsection{Binary Classification}

\subsubsection{Breast Cancer Wisconsin (Diagnostic) Dataset [\cite{8}]}

The binary classification model is applied to the Breast Cancer Wisconsin (Diagnostic) Dataset [\cite{8}], utilizing a single-layer gated perceptron with one neuron. The inputs to the gated perceptron are $n$ entries $X_i$, and the output $y$ is computed as the sigmoid of the sum of weighted inputs (see Figure \ref{fig10}). Additionally, the product $(X_1, X_2, ..., X_n)$,
 is introduced as a new term in the output expression. This product is computed once and treated as a weighted input.

\begin{equation*}
Sum= \omega_1 X_1 + \omega_2 X_2 +..+ \omega_n X_n + \omega_{n+1} X_1 X_2 .. X_n + b
\end{equation*}
\begin{equation*}
y=sigmoid(Sum)
\end{equation*}

\begin{figure}[ht!]
\centering
\includegraphics[width=8cm]{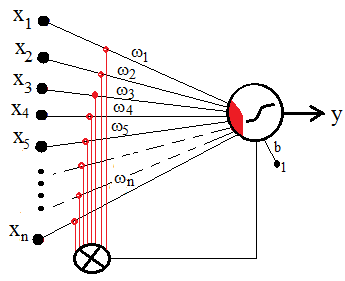}
\caption{\label{fig10} The gated perceptron with $n$ inputs used for binary classification.}
\end{figure}

The gated perceptron uses a Sigmoid activation function to map inputs to a probability value between 0 and 1, which is then interpreted as either class 0 (benign) or class 1 (malignant). The weights of the model are initialized, and the model attempts to learn optimal values through training.

The dataset used is breast cancer data, with $32$ columns, where the 'Diagnosis' column is mapped to 1 for malignant (M) and 0 for benign (B) diagnoses. The data is normalized to ensure that all feature values are scaled between 0 and 1. The data is split into training and testing sets, with 80\% of the data used for training and 20\% for testing.

The training is done over multiple epochs, in each epoch, the model computes the error between the true label and the predicted output and updates the weights based on the gradient of the error using back propagation and the sigmoid derivative. 

The model's performance is evaluated using common classification metrics: Accuracy, Precision, Recall, F1 Score. Additionally, the binary cross-entropy loss is computed and stored for each epoch to track the model’s performance over time. After training, the model is evaluated on the test data using the same metrics. The ROC-AUC score and ROC curve is also be computed to evaluate the model's ability to distinguish between the two classes.

\begin{figure}[ht!]
\centering
\includegraphics[width=10cm]{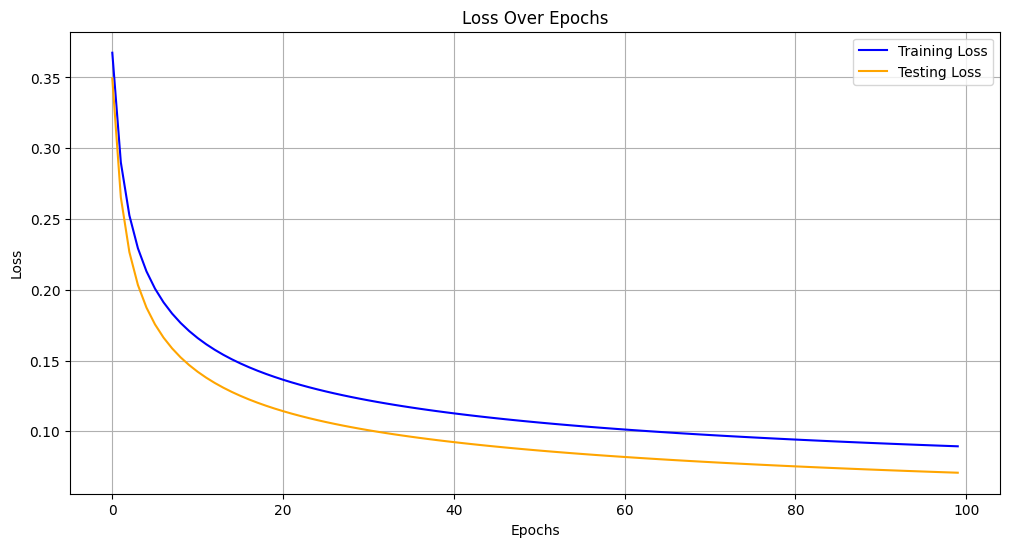}
\caption{\label{res1} The loss function for the gated perceptron applied to wdbc dataset.}
\end{figure}

\begin{figure}[ht!]
\centering
\includegraphics[width=14cm]{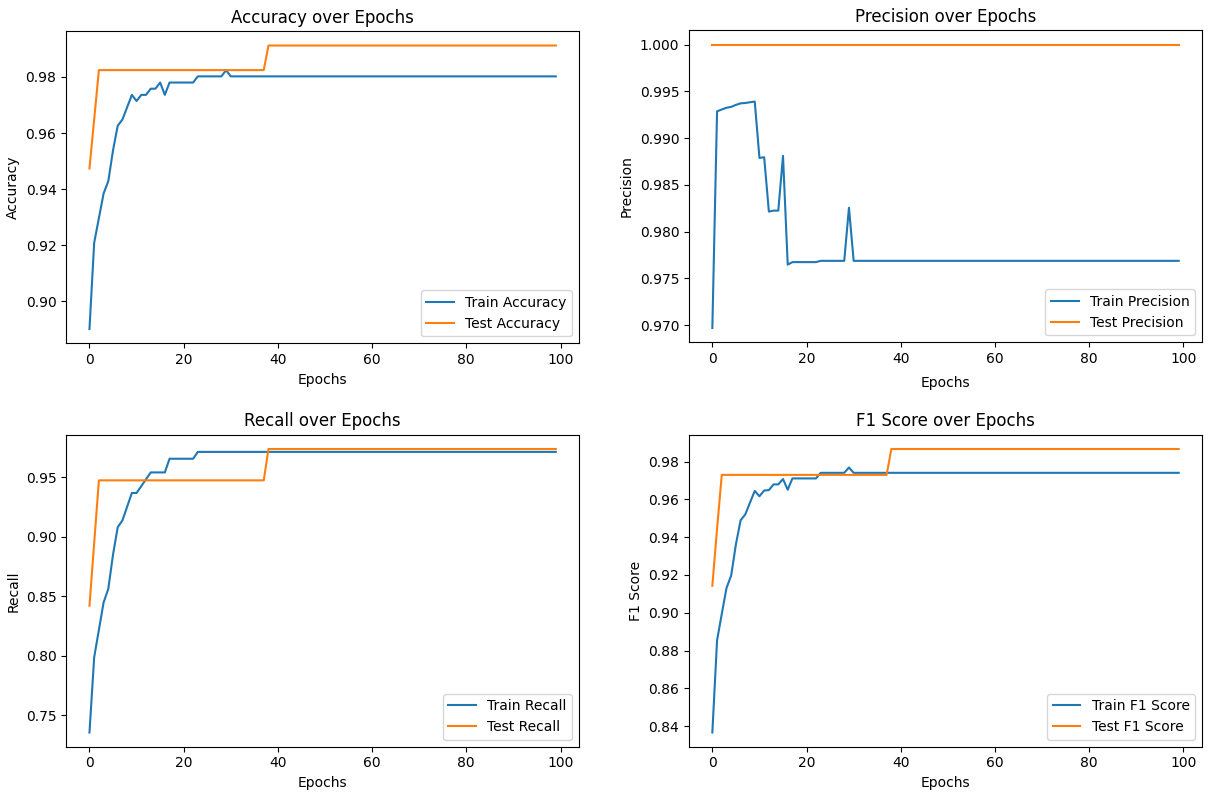}
\caption{\label{res2} The accuracy, precision, recall, F1Score curves for the gated perceptron applied to wdbc dataset.}
\end{figure}

\begin{figure}[ht!]
\centering
\includegraphics[width=10cm]{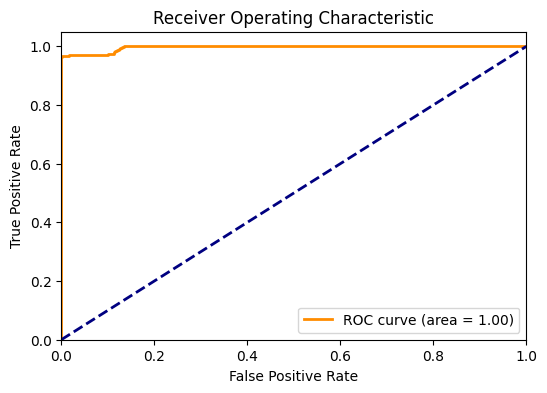}
\caption{\label{res3} The AUC curve for the gated perceptron applied to wdbc dataset.}
\end{figure}

The results obtained with a learning rate of 0.5 and 100 epochs are presented in Table \ref{tab1}, showing values for True Positive (TP), True Negative (TN), False Positive (FP), False Negative (FN), Accuracy (AC), Precision (Pr), Recall (Rec), F1 Score (F1), and Area Under the Curve (AUC) from ten successive runs of our code [\cite{9}] on randomly chosen test data. Note that learning rates in the range of 0.1 to 1.0 yield similar results. Learning rates outside this interval, such as 0.05, 0.01, 1.2, or 1.3, also produce comparable outcomes. Convergence of the system typically occurs around 60 epochs. With a single gated perceptron, our results are competitive with state-of-the-art methods and achieve very low values for False Positives $(0.7)$ and False Negatives $(1.5)$. Figures \ref{res1}, \ref{res2}, \ref{res3} illustrate the graphs corresponding to the different measures.

\begin{table}
    \centering
    \begin{tabular}{ccccccccc}
        TP & TN & FP & FN & Ac & Pr & Rec & F1 & AUC \\
        \hline
        39 & 73 & 1 & 1 & 0.982 & 0.975 & 0.975 & 0.975 & 0.999\\
         43 & 70 & 0 & 1 & 0.991 & 1.0  & 0.977 & 0.988 & 0.999\\
         44 & 67 & 2 & 1 & 0.974 & 0.956 & 0.978 & 0.967 & 0.997\\
         46 & 65 & 0 & 3 & 0.974 & 1.0 & 0.939 & 0.968 & 0.996 \\
         46 & 66 & 1 & 1 & 0.982 & 0.979 & 0.979 & 0.979 & 0.997 \\
         40 & 72 & 0 & 2 & 0.982 & 1.0 & 0.952 & 0.976 & 0.995 \\
         38 & 74 & 0 & 2 & 0.982 & 1.0 & 0.950 & 0.974 & 0.999 \\
         40 & 71 & 1 & 2 & 0.977 & 0.976 & 0.952 & 0.964 & 0.994 \\
         29 & 84 & 0 & 1 & 0.991 & 1.0 & 0.967 & 0.983 & 0.980 \\
         53 & 58 & 2 & 1 & 0.974 & 0.964 & 0.981 & 0.972 & 0.998 \\
           \textbf{Mean}&   &  \textbf{0.7} & \textbf{1.5} & \textbf{0.98} & \textbf{0.985} & \textbf{0.965} & \textbf{0.975} & \textbf{0.995}
    \end{tabular}
    \caption{The different values of measure obtained for 10 successive run of the code with a gated perceptron.}
    \label{tab1}
\end{table}


\subsubsection{Discussion}

To understand the good results obtained with only one gated perceptron, we tracked the values of the weights associated with the added input (the gate) across all epochs. The weights remained stable, indicating that the gated perceptron performed computations similar to a traditional perceptron.

\begin{figure}[ht!]
\centering
\includegraphics[width=10cm]{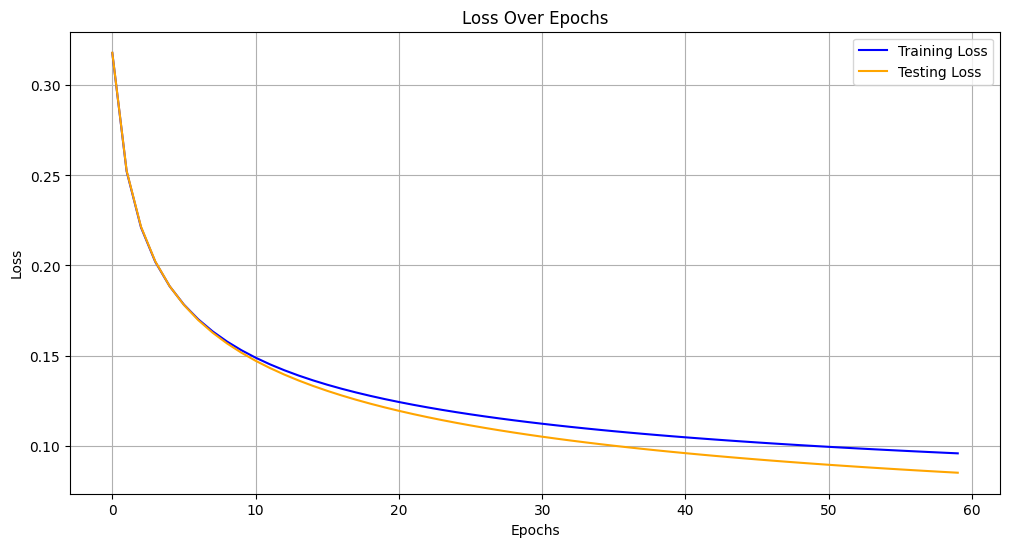}
\caption{\label{res4} The loss function for the perceptron applied to wdbc dataset.}
\end{figure}

\begin{figure}[ht!]
\centering
\includegraphics[width=14cm]{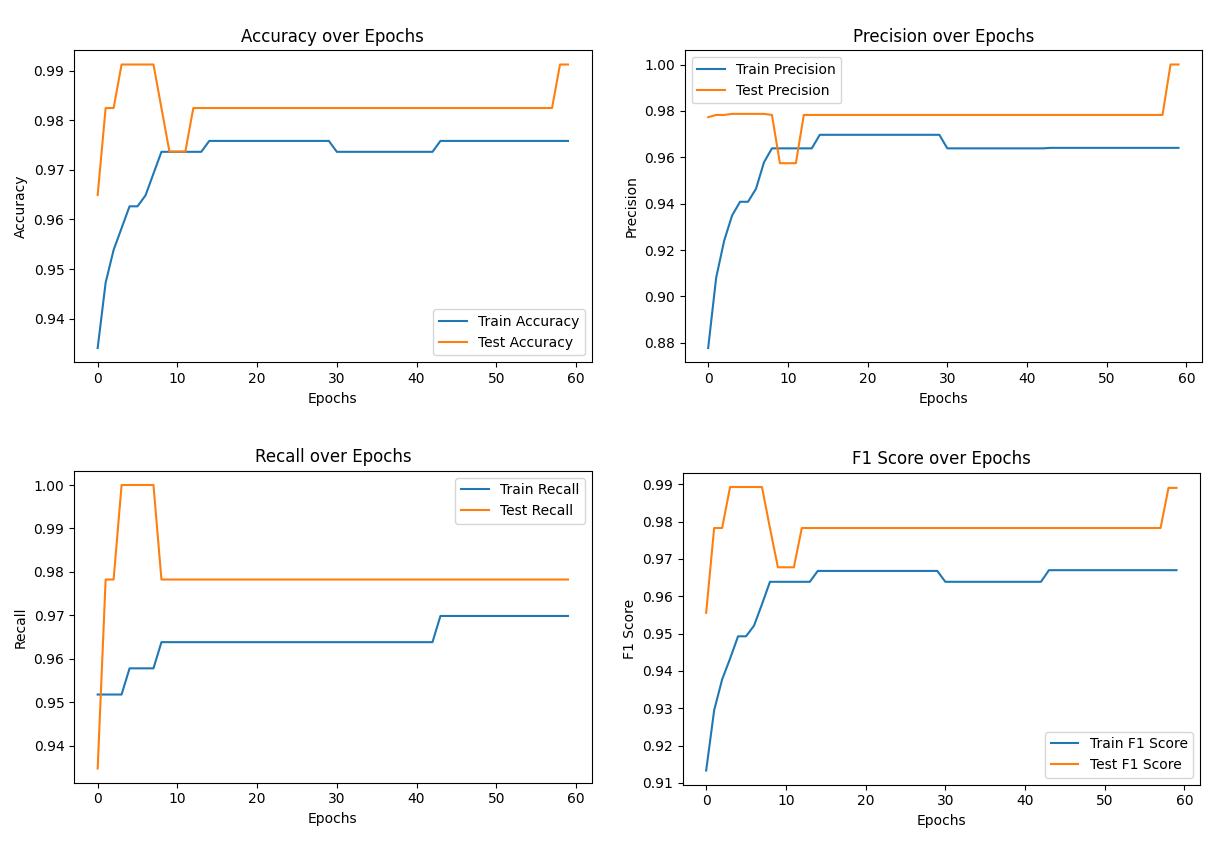}
\caption{\label{res5} The accuracy, precision, recall, F1Score curves for the perceptron applied to wdbc dataset.}
\end{figure}

\begin{figure}[ht!]
\centering
\includegraphics[width=10cm]{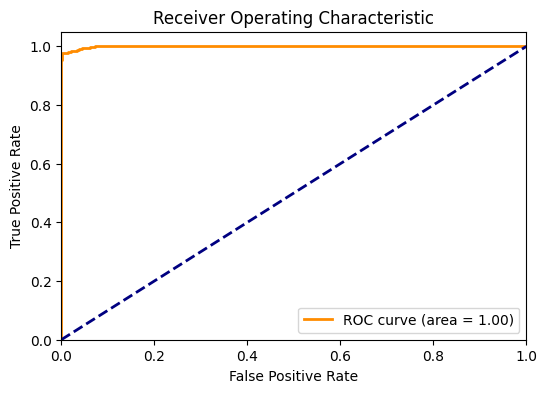}
\caption{\label{res6} The AUC curve for the perceptron applied to wdbc dataset.}
\end{figure}

When we replaced the gated perceptron with a conventional perceptron, we obtained the same results (see Table \ref{tab2} and Figures \ref{res4}, \ref{res5}, \ref{res6}). This suggests that the 30 features of the WDBC (Wisconsin Diagnostic Breast Cancer) dataset are effectively linear. This finding is noteworthy because many researchers have developed various methods, including complex neural networks, without testing with a single perceptron, under the assumption that the WDBC dataset is not linear.
Indeed, the dataset comprises various measurements of cell nuclei, and non-linearity is expected because features related to complex biological systems are often highly non-linear. Interactions between features (e.g., how radius, texture, and smoothness collectively predict malignancy) are generally not just linear. We made our code publicly available on GitHub for testing [\cite{9}].

\begin{table}
    \centering
    \begin{tabular}{ccccccccc}
        TP & TN & FP & FN & Ac & Pr & Rec & F1 & AUC \\
        \hline
        36 & 76 & 0 & 2 & 0.982 & 1.0 & 0.947 & 0.972 & 0.999\\
        44 & 68 & 0 & 2 & 0.997 & 0.982 & 1.0 & 0.956 & 0.997\\
        45 & 65 & 0 & 4 & 0.965 & 1.0  & 0.918  & 0.957 & 0.987\\
        42 & 68 & 0 & 4 & 0.965 & 1.0 & 0.913 & 0.955 & 0.994\\
        42 & 68 & 1 & 3 & 0.965 & 0.977 & 0.933 & 0.955 & 0.994\\
        38 & 75 & 0 & 1 & 0.991 & 1.0 & 0.974 & 0.987 & 0.999\\
        45 & 67 & 1 & 1 & 0.982 & 0.978 & 0.978 & 0.978 & 0.999\\
        39 & 74 & 0 & 1 & 0.991 & 1.0 & 0.975 & 0.987 & 0.998\\
        40 & 71 & 0 & 3 & 0.973 & 1.0 & 0.930 & 0.963 & 0.999\\
        49 & 61 & 2 & 2 & 0.964 & 0.960 & 0.960 & 0.960 & 0.990\\
           \textbf{Mean}&   &  \textbf{0.4} & \textbf{2.3} & \textbf{0.977} & \textbf{0.989} & \textbf{0.953} & \textbf{0.967} & \textbf{0.995}
    \end{tabular}
    \caption{The different values of measure obtained for 10 successive run of the code related to wdbc dataset with a perceptron.}
    \label{tab2}
\end{table}


The obtained results compete those obtained in published methods so far. We can find the most relevant methods and scores in [\cite{12}]. The classifiers Support vector machine (SVM), Random Forest (RF), K-nearest neighbors(K-NN), Decision tree (DT), Naïve Bayes (NB), Logistic
Regression (LR), AdaBoost (AB), Gradient Boosting (GB), Multi-layer perceptron (MLP), Nearest Cluster Classifier (NCC), and voting classifier (VC) have been used for comparing and analyzing breast cancer into benign and malignant tumors. the result shows that the Voting classifier has the highest accuracy, which is 98.77\%, with the lowest error rate. The results are given by table \ref{tabref1}.

\begin{table}
    \centering
    \begin{tabular}{cccccc}
        Methods & Precision & Recall & $F_Measure$ & Accuracy & Error\\
        \hline
        SVM & 98.28 \% & 97.61 \% & 97.92 \% &  98.07 \% &  0.019 \% \\
        RF & 93.91 \% &  93.65 \% &  93.78 \% &  94.20 \% &  0.058 \% \\ 
        KNN & 97.22 \% &  96.04 \% &  96.58 \% &  96.84 \% &  0.061 \% \\ 
        DT & 93.99 \% &  93.56 \% &  93.77 \% &  94.20 \% &  0.058 \% \\ 
        NB & 90.73 \% &  89.98 \% &  90.33 \% &  91.04 \% &  0.089 \% \\ 
        LR & 98.55 \% &  98.07 \% &  98.30 \% &  98.42 \% &  0.015 \% \\ 
        AB & 96.27 \% &  95.81 \% &  96.03 \% &  96.31 \% &  0.036 \% \\ 
        GB & 95.57 \% &  95.39 \% &  95.48 \% &  95.78 \% &  0.042 \% \\ 
        MLP & 97.55 \% &  97.18 \% &  97.36 \% &  97.54 \% &  0.024 \% \\ 
        NCC & 93.44 \% &  91.86 \% &  92.54 \% &  93.15 \% &  0.068 \% \\ 
        VC (LR+SVM) & 98.83 \% &  98.54 \% &  98.68 \% &  98.77 \% &  0.012 \% \\ 
    \end{tabular}
    \caption{Evaluation of classification methods after feature optimization.}
    \label{tabref1}
\end{table}

\subsubsection{PIMA Indian Dataset [\cite{7}]}

We implemented a single-layer gated perceptron model to classify patients as diabetic or non-diabetic based on the PIMA Indian Dataset [\cite{7}] using one gated perceptron.

We performed data preprocessing to ensure that missing values were handled appropriately, and we normalized all features. The model was trained using gradient descent with sigmoid activation and binary cross-entropy loss, and validated on a separate test set using various performance metrics.

The same experiment has been conducted using a mode with one perceptron. The results are given by tables \ref{tab3gp} and \ref{tab3p}.
Globally, both the gated and conventional perceptron achieve similar results. Note that the gated perceptron performs better overall considering the F1 Score and Recall while maintaining reasonable precision and overall performance which are critical in medical contexts because minimizing false negatives is more critical.

\begin{table}
    \centering
    \begin{tabular}{ccccccccccc}
        TP & TN & FP & FN & Ac & Pr & Rec & F1 & AUC \\
        \hline
 
        & 29 & 87 & 17 & 21 & 0.753 & 0.630 & 0.58 & 0.604 & 0.826 \\
        & 41 & 82 & 13 & 18 & 0.799 & 0.759 & 0.695 & 0.726 & 0.877\\
        & 26 & 81 & 18 & 29 & 0.695 & 0.591 & 0.473 & 0.525 & 0.792 \\
        & 37 & 79 & 20 & 18 & 0.753 & 0.649 & 0.673 & 0.661 & 0.809 \\
        & 33 & 78 & 21 & 22 & 0.721 & 0.611 & 0.6 & 0.606 & 0.799 \\
        & 32 & 87 & 22 & 13 & 0.774 & 0.593 & 0.711 & 0.646 & 0.839 \\
        & 34 & 81 & 14 & 25 & 0.747 & 0.708 & 0.576 & 0.636 & 0.809 \\
        & 26 & 89 & 19 & 20 & 0.747 & 0.578 & 0.565 & 0.571 & 0.799 \\
        & 33 & 85 & 21 & 15 & 0.766 & 0.611 & 0.687 & 0.647 & 0.846 \\
        & 31 & 93 & 6 & 24 & 0.805 & 0.838 & 0.564 & 0.674 & 0.853 \\
     \textbf{Mean} & \textbf{32}& \textbf{84}  &  \textbf{17} & \textbf{20} & \textbf{0.756} & \textbf{0.659} & \textbf{0.612} & \textbf{0.630 } & \textbf{0.825}
    \end{tabular}
    \caption{The different values of measure obtained for 10 successive run of the code related to diabetes dataset with one gated perceptron.}
    \label{tab3gp}
\end{table}

\begin{table}
    \centering
    \begin{tabular}{ccccccccccc}
        TP & TN & FP & FN & Ac & Pr & Rec & F1 & AUC \\
        \hline
    & 25 & 90 & 14 & 25 & 0.747 & 0.641 & 0.5 & 0.562  & 0.768 \\
    & 39 & 79 & 20 & 16 & 0.766 & 0.661 & 0.709 & 0.684 & 0.843 \\
    & 24 & 89 & 2 & 39 & 0.734 & 0.923 & 0.381 & 0.539 & 0.886 \\
    & 29 & 88 & 14 & 23 & 0.759 & 0.674 & 0.557 & 0.611 & 0.823 \\
    & 20 & 89 & 2 & 43 & 0.708 & 0.909 & 0.317 & 0.470 & 0.861 \\
    & 21 & 100 & 10 & 23 & 0.786 & 0.677 & 0.477 & 0.56 & 0.833 \\
    & 30 & 90 & 12 & 22 & 0.779 & 0.714 & 0.577 & 0.638 & 0.822 \\
    & 37 & 82 & 19 & 16 & 0.772 & 0.661 & 0.698 & 0.679 & 0.840 \\
    & 23 & 94 & 11 & 26 & 0.759 & 0.676 & 0.469 & 0.554 & 0.823 \\
    & 31 & 83 & 12 & 28 & 0.740 & 0.721 & 0.525 & 0.608 & 0.818\\
     \textbf{Mean} & \textbf{28} &  \textbf{88} &  \textbf{12} & \textbf{26} & \textbf{0.755} & \textbf{0.725} & \textbf{0.521} & \textbf{0.590} & \textbf{0.831}
    \end{tabular}
    \caption{The different values of measure obtained for 10 successive run of the code related to diabetes dataset with one perceptron.}
    \label{tab3p}
\end{table}

The obtained results compete those obtained in published methods so far. We can cite the most relevant methods and scores in the table \ref{tabref2} [\cite{13}].

\begin{table}
    \centering
    \begin{tabular}{cccccc}
       Algorithm & Accuracy & Precision & Recall & F1 score & AUC\\
       \hline
          &   &   &  Without PCA & &  \\ 
        SVM & 79.02 & 74.43 & 78.03 & 71.33 & 87.22\\
        NB(Naive Bayes) & 78.18 & 73.43 & 78.92 & 76.53 & 87.87 \\
        RF (Random Forest) & 83.65& 86.98 & 75.65 & 80.02 & 77.94 \\
        DT(Decision Trees) & 72.55 & 71.12 & 73.04 & 72.01 & 80.82 \\
          &  &  &  With PCA &  & \\
        SVM & 86.08 & 88.88 & 86.90 & 88.65 & 92.91
  \end{tabular}
    \caption{All Model Performance for the 80\% and 20\% of training and testing ratio..}
    \label{tabref2}
\end{table}

\subsection{Multi-Class Classification}

We performed multi-class classification on the Iris dataset [\cite{11}] using a single-layer gated perceptron model with softmax output for the three classes: Iris-setosa, Iris-versicolor, and Iris-virginica.

The Iris dataset was preprocessed by mapping the class labels ('type') to integers as follows: Iris-setosa → 0, Iris-versicolor → 1, and Iris-virginica → 2. A new feature, referred to as 'product,' was introduced by calculating the product of the four input features (x1, x2, x3, x4). Each feature, including the 'product' column, was normalized to a range between 0 and 1. The dataset was then split into training and test sets using an 80-20 split.

The gated perceptron's output was computed using the softmax function, converting raw logits into probabilities for each class. The model was trained using gradient descent, where the error was calculated as the difference between the predicted and true labels (one-hot encoded). The model’s weights were updated using a learning rate of 0.01 based on the error.

Finally, we computed the confusion matrix, which revealed how well the model predicted each class.

The following test accuracy scores were obtained by running the model 10 times successively with a random selection of test data: 1.0000, 0.9333, 0.9667, 0.9667, 0.9000, 0.9667, 0.9667, 0.9333, 0.9667, and 0.9000. These results yield an average accuracy of 0.950.

Comparing with the state of art methods, the average accuracy rates (\%) obtained by the models: MNBHL, AdaBoost, Bagging of MLP, Decision Tree, Logistic Regression, MLP, Naive-Bayes, Random Forest and SVM are 96.5, 95.2, 97.8, 95.6, 94.3, 96.9, 96.5, 95.6, 97.8 [\cite{11}].

\section{Conclusion}

In this paper, we introduced the gated perceptron as an enhancement over the conventional perceptron, allowing it to handle non-linearity in data through the introduction of a new input that captures interactions between features. We demonstrated how the gated perceptron can generate more distinct regions in the input space, improving its ability to perform both linear and non-linear regression and classification tasks.

Our experiments, conducted on both binary and multi-class classification problems, as well as regression tasks using common datasets like Iris and Breast Cancer Wisconsin, illustrate the benefits of using a gated perceptron. Notably, the gated perceptron outperformed the conventional perceptron in scenarios requiring non-linear decision boundaries, particularly in handling the complex datasets.

The results show that the gated perceptron is competitive with state-of-the-art methods for classification and regression, while maintaining simplicity in its architecture. This makes it a promising tool for applications where interpretability and performance are crucial. Future work could extend the use of gated perceptrons in deeper neural networks and explore its application in more complex data structures.


\end{document}